\documentclass{article}

    \usepackage[preprint]{neurips_2025}

\usepackage[utf8]{inputenc} 
\usepackage[T1]{fontenc}    
\usepackage{hyperref}       
\usepackage{url}            
\usepackage{booktabs}       
\usepackage{amsfonts}       
\usepackage{nicefrac}       
\usepackage{microtype}      
\usepackage{xcolor}         
\usepackage{graphicx}
\usepackage{multirow} 
\usepackage{natbib}
\usepackage{enumitem}
\usepackage{amsmath}
\usepackage{tabularx}
\usepackage{subfigure}
\usepackage{subcaption}

\title{RoboScape: Physics-informed Embodied World Model}

\author{
Yu Shang$^{1}$~~~~Xin Zhang$^{2}$~~~~Yinzhou Tang$^1$~~~~Lei Jin$^{1}$~~~~Chen Gao$^{1}$~~~~\textbf{Wei Wu}$^{2}$\footnotemark[1]~~~~\textbf{Yong Li}$^1$\footnotemark[1]   \\
\\
~~~$^1$Tsinghua University \\
~~~$^2$Manifold AI
}

\begin{document}

\maketitle
\footnotetext[1]{Corresponding author, correspondence to liyong07@tsinghua.edu.cn.}

\begin{abstract}
World models have become indispensable tools for embodied intelligence, serving as powerful simulators capable of generating realistic robotic videos while addressing critical data scarcity challenges. However, current embodied world models exhibit limited physical awareness, particularly in modeling 3D geometry and motion dynamics, resulting in unrealistic video generation for contact-rich robotic scenarios. In this paper, we present RoboScape, a unified physics-informed world model that jointly learns RGB video generation and physics knowledge within an integrated framework. We introduce two key physics-informed joint training tasks: temporal depth prediction that enhances 3D geometric consistency in video rendering, and keypoint dynamics learning that implicitly encodes physical properties (e.g., object shape and material characteristics) while improving complex motion modeling.
Extensive experiments demonstrate that RoboScape generates videos with superior visual fidelity and physical plausibility across diverse robotic scenarios. We further validate its practical utility through downstream applications including robotic policy training with generated data and policy evaluation.
Our work provides new insights for building efficient physics-informed world models to advance embodied intelligence research.
The code is available at: \url{https://github.com/tsinghua-fib-lab/RoboScape}.

\end{abstract}

\section{Introduction}
The advancement of large language and vision models~\cite{achiam2023gpt,team2023gemini} has demonstrated the critical role of high-quality, large-scale training data for robust generalization. 
However, the robotic learning is significantly hindered by the prohibitive cost of collecting real-world data~\cite{walke2023bridgedata,o2024open,khazatsky2024droid,bu2025agibot}, which often relies on human teleoperation to acquire high-quality demonstrations.
This limitation poses a great challenge for scaling robotic learning and deploying agents in complex, real-world environments.

World models~\cite{ha2018recurrent,yang2023learning,brooks2024video}, which simulate environmental dynamics by predicting future states based on current observations and given actions, offer a promising solution to this data scarcity problem.
Such models hold significant promise for advancing embodied intelligence by generating realistic robotic data~\cite{wang2025learning} and enabling scalable simulation environments~\cite{wu2024ivideogpt}.
However, current embodied world models~\cite{wu2024ivideogpt,zhu2024irasim,agarwal2025cosmos} predominantly focus on video generation, with training objectives centered on optimizing the RGB pixels. 
While capable of producing visually plausible 2D images, they often fail to maintain crucial physical properties, such as motion plausibility and spatial consistency~\cite{kang2024far}. 
Particularly, in robotic manipulation tasks involving deformable objects (e.g., cloth), the generated videos frequently contain artifacts such as unrealistic object morphing or discontinuous motion.
These limitations become particularly detrimental in interaction-rich robotic scenarios, where even minor physical inconsistencies can dramatically compromise the effectiveness of learned policies.

The root cause lies in existing models’ overreliance on visual token fitting without awareness of physical knowledge~\cite{lin2025exploring,guo2025t2vphysbench,meng2024towards}. 
To address this, we propose a physics-informed world model that jointly learns depth information and temporal keypoint consistency to implicitly encode physical constraints. 
Existing efforts of integrating physical knowledge into video generation fall into three categories: physics-prior regularization, physics simulator-based knowledge distillation, and material field modeling. Current regularization-based methods enforce constraints such as local rigidity~\cite{chen2025towards} or rotational similarity~\cite{rai2024enhancing} on Gaussian splatting (GS) features or 3D point clouds. 
However, these methods are limited to narrow domains like human motion~\cite{luiten2024dynamic} or rigid-body dynamics~\cite{chen2025towards}, hindering generalization to diverse robotic scenarios.
Another line of work employs physics simulators to extract motion signals or semantic maps as conditions to guide video generation models~\cite{xie2025physanimator,lv2024gpt4motion,liu2024physgen,tan2024physmotion}. 
Although this approach yields reliable physical priors, the resulting cascaded pipeline introduces excessive computational complexity, hindering their practical deployment.
There have been some recent works trying to enhance the physical simulation via material field modeling~\cite{zhang2024physdreamer,liu2024physics3d}.
However, such methods are confined to object-level modeling and are hard to apply to scene-level generation.

To overcome these limitations, this work addresses a fundamental challenge: how to effectively integrate physical knowledge into world model learning in a unified and computationally efficient framework, eliminating the need for complex model cascades or additional training pipelines.
We propose RoboScape, a physics-informed world model based on a multi-task learning auto-regressive framework to generate visually realistic and physics-adherent robotic videos, effectively controlled by robot actions and current observations. 
Specifically, our approach incorporates physics knowledge through two auxiliary physics-informed supervision tasks within the world model itself to alleviate heavy external model cascading. 
First, to empower the model with 3D spatial physical understanding, we augment the RGB prediction backbone with a temporal depth prediction branch and inject the learned depth features into the RGB prediction to enhance spatial awareness. 
Such synergistic learning of temporal depth maps enables the model to implicitly acquire 3D scene reconstruction priors rather than merely fitting 2D RGB images. 
Second, we introduce an adaptive keypoint dynamics learning task to address unrealistic object deformation and implausible motion issues. 
To achieve this, we first perform dynamic keypoint sampling to automatically identify regions with significant motion (typically involving robots and interacting objects), then enforce temporal token consistency for these keypoints across frames. 
Through this, the model effectively captures the deformation properties and motion behaviors of objects, implicitly encoding material properties (e.g., rigidity and softness) through self-supervised keypoint consistency, eliminating the need for explicit material modeling.
Although some recent world models~\cite{team2025aether,zhen2025tesseract} also explore joint RGB-depth prediction, their learning remains constrained at the whole image level, failing to capture the fine-grained motion dynamics and object deformation details that are crucial for robotic manipulation scenarios.
Furthermore, these approaches exhibit a performance trade-off, where gains in 3D perception come at the cost of reduced RGB prediction fidelity.
Differently, our model offers a more comprehensive solution that captures global spatial knowledge through learning temporal depth dynamics, while modeling local object deformation and motion characteristics via learning temporal keypoint tracking. 

We conduct comprehensive experiments to evaluate our world model from three aspects: video generation quality, robotic policy learning using synthetic data, and robotic policy evaluation. 
RoboScape achieves state-of-the-art performance in both RGB and depth prediction accuracy, achieving a superior balance between these metrics compared to existing world model baselines. 
Additionally, we validated that using synthetic data from our world model consistently improves the performance of classic robotic policy models including Diffusion Policy~\cite{chi2023diffusion} and pi0~\cite{black2024pi_0}, confirming the model's practical utility for robotic learning. 
Finally, our model can also serve as a reliable policy evaluator, with assessment results showing strong correlation with ground-truth simulator outcomes, confirming our model's capability to accurately model the physical world.

In summary, the main contributions of the paper are as follows:
\begin{itemize}[leftmargin=*]
\item We propose RoboScape, a physics-informed embodied world model that unifies RGB video generation, temporal depth prediction, and adaptive keypoint tracking in a joint learning framework, achieving both high visual fidelity and physical plausibility.
\item We design an automated robotic data processing pipeline with physical prior information labels. Trained on the carefully curated large-scale, high-quality dataset, our model achieves SOTA performance on visual quality, geometric accuracy, and action controllability.
\item We demonstrate the practical utility of RoboScape on downstream applications including robotic policy training and evaluation. Extensive experimental results demonstrate its effectiveness in accurately modeling embodied environments, validating its potential for advancing real-world robotic deployment.
\end{itemize}

\section{Methodology}
\subsection{Problem Formulation}
In this work, we focus on robot manipulation scenarios and learn an embodied world model $f_\theta$ as a dynamics function that predicts the next visual observation $\mathbf{o}_{t+1}$ given past observations $\mathbf{o}_{1:t}$ and robotic actions $\mathbf{a}_{1:t}$:
\begin{equation}
    \mathbf{o}_{t+1} \sim f_{\theta}(\mathbf{o}_{t+1} | \mathbf{o}_{1:t}, \mathbf{a}_{1:t}),
\end{equation}
where $\mathbf{o} \in \mathbb{R}^{H \times W \times 3}$ is a video frame and $\mathbf{a} \in \mathbb{R}^k$ is a $k$-degree continuous action control vector.

\subsection{Robotic Data Processing Pipeline with Physical Priors Annotation}
\begin{figure}[t]
    \centering
    \includegraphics[width=0.99\linewidth]{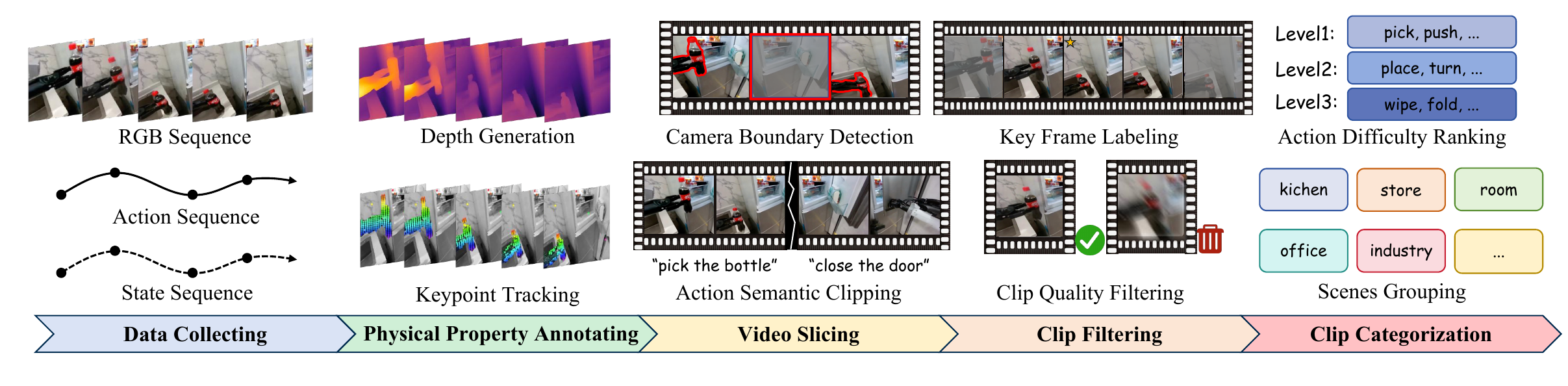}
    \caption{Illustration of the proposed robotic data processing pipeline with physical priors annotation.}
    \label{fig:data}
\end{figure}
Learning a physics-informed embodied world model requires high-quality dataset covering high-resolution RGB and depth sequences, action sequences that control the robot, and state sequences that the robot executes. In this section, we present our data processing pipeline to construct a multi-modal embodied dataset with physical priors based on AGIBOT-World dataset~\cite{bu2025agibot}, as shown in Fig.~\ref{fig:data}.

\textbf{Physical Property Annotating based on Depth Generation and Keypoint Tracking.}
We focus on two fundamental physical priors in videos: temporal depth consistency and keypoint motion trajectories. These features can be efficiently extracted using off-the-shelf pretrained models, enabling enhanced generalization while maintaining practical feasibility.
Specifically, we utilize Video Depth Anything~\cite{chen2025video} to generate the depth map sequence of the video. Furthermore, we apply SpatialTracker~\cite{xiao2024spatialtracker} as the keypoint tracking model to sample the keypoint and track their trajectories.

\textbf{Video Slicing based on Camera Boundary Detection and Action Semantic.}
The original videos have different attributes, such as lengths and resolution, with camera jumps or editing traces, and a video may contain multiple action semantics. Thus, we slice the video into clips with normalized attributes, consistent motion, no camera jumps, and single action semantics. Specifically, we use TransNetV2~\cite{soucek2024transnet} to perform camera boundary detection and use Intern-VL~\cite{chen2024internvl} to generate the action semantic of a specific clip.

\textbf{Clip Filtering based on Key Frame and Clip Quality.}
The generated clips are highly heterogeneous in terms of quality, semantics, and presentation form. To ensure the validity and adaptability of the training data, we introduce a clip filtering mechanism including: (1) using FlowNet~\cite{dosovitskiy2015flownet} to filter out clips with indistinct motion and disordered movement patterns, and (2) using Intern-VL~\cite{chen2024internvl} to label the key frame of the clip and filter out the frames without explicit relationship to the key frame.

\textbf{Clip Categorization based on Action Difficulty and Scenes.}
In this stage, we categorize and reorganize the dataset based on action difficulty and clip scenes to support the curriculum learning strategy~\cite{wang2021survey}, which trains the world model from easier to harder tasks.

\subsection{RoboScape: A Physics-informed Embodied World Model}
\begin{figure}[t]
    \centering
    \includegraphics[width=0.98\linewidth]{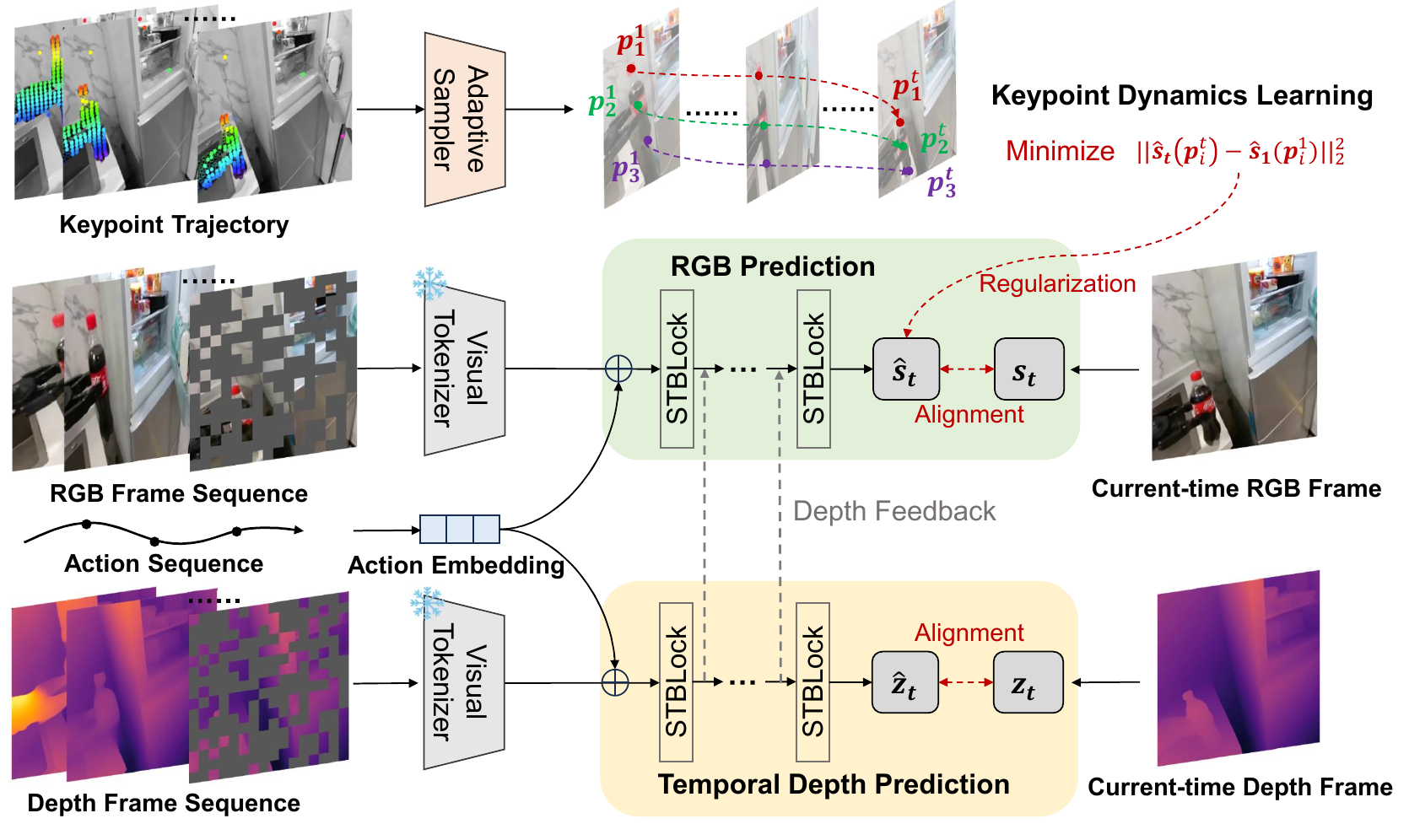}
    \caption{Overview of the physics-informed world model, where physical knowledge is integrated through joint learning of temporal depth estimation and adaptively sampled keypoint dynamics.}
    \label{fig:model}
\end{figure}

RoboScape is designed to achieve frame-level action-controllable robot video generation, enabling interactive future frame prediction. 
At its core, we adopt an auto-regressive Transformer-based framework that iteratively predicts the next frame based on historical frames and the current robot action. 
To enhance the physical plausibility of generated videos, we introduce two physics-informed auxiliary training tasks in addition to the normal RGB image prediction: (1) temporal depth prediction, which enforces global geometric consistency across frames, and (2) adaptively sampled keypoint dynamics learning, which captures the motion and deformation details of local dynamic objects. 
The whole pipeline is illustrated in Figure~\ref{fig:model}.
Joint training with these physics-aware regularizers provides an efficient approach to embed physical priors into world models, significantly reducing the reality gap between generated videos and real-world dynamics.

\textbf{Video Tokenization.}
To enable efficient video generation, we leverage MAGVIT-2~\cite{yu2023language} to compress raw RGB frames $\mathbf{o}_{1:T} \in \mathbb{R}^{T \times H \times W \times 3}$ into discrete latent tokens $\mathbf{s}_{1:T} \in \mathbb{R}^{T \times H' \times W' \times D}$, where $H' = H/\alpha$ and $W' = W/\alpha$ denote the reduced spatial dimensions ($\alpha$ being the downsampling factor), and $D$ represents the latent channel dimension. Similarly, we tokenize temporal depth maps $\mathbf{d}_{1:T} \in \mathbb{R}^{T \times H \times W \times 1}$ into latent depth tokens $\mathbf{z}_{1:T} \in \mathbb{R}^{T \times H' \times W' \times D}$.

\textbf{Geometry Consistency Enhancement via Temporal Depth Prediction.}
While RGB-based video generation has achieved remarkable progress, it often suffers from inconsistent 3D geometry due to the lack of explicit spatial constraints. 
Considering that inter-frame depth variations encode crucial 3D structure information, we propose to jointly learn temporal RGB and depth information, leveraging depth features as geometric constraints to ensure spatially coherent video generation.
For joint prediction of both RGB and depth images, we propose a dual-branch co-autoregressive Transformer (DCT). 
Each branch consists of stacked Spatial-Temporal Transformer (ST-Transformer) blocks, which implement a causal attention mechanism in the temporal attention layers for generation causality, and bidirectional attention in the spatial attention layers to enable full context modeling. 

At timestep $t$, the model processes historical latent tokens through parallel branches $\mathcal{F}_{\text{RGB}}$ and $\mathcal{F}_{\text{Depth}}$, conditioned on learned action embeddings $\mathbf{c}_{1:t-1} \in \mathbb{R}^{(t-1) \times 1 \times 1 \times D} : \mathcal{E}_a(\mathbf{a}_{1:t-1})$ and position embeddings $\mathbf{e}_{1:t-1} \in \mathbb{R}^{(t-1) \times H' \times W' \times D}$, where $\mathcal{E}_a$ denotes the robot action encoder. The auto-regressive prediction of each branch is formulated as:
\begin{equation}
\begin{aligned}
\hat{\mathbf{s}}_t &= \mathcal{F}_{\text{RGB}}(\mathbf{s}_{1:t-1} \oplus \mathbf{c}_{1:t-1} \oplus \mathbf{e}_{1:t-1}), \\
\hat{\mathbf{z}}_t &= \mathcal{F}_{\text{Depth}}(\mathbf{z}_{1:t-1} \oplus \mathbf{c}_{1:t-1} \oplus \mathbf{e}_{1:t-1}),
\end{aligned}
\end{equation}
where $\oplus$ denotes element-wise addition with broadcasting.
Empirically, we find that simple additive fusion provides effective action control while maintaining model efficiency.

To inject depth predictions as physical priors into the RGB branch and enhance spatial structure fidelity of rendered videos, we introduce cross-branch interaction pathways. 
Specifically, at each ST-Transformer block $l$, we project the depth branch's intermediate features $\mathbf{h}_{\text{depth}}^l$ and fuse them additively with the corresponding RGB features:
\begin{equation}
\mathbf{h}_{\text{RGB}}^l = \mathbf{h}_{\text{RGB}}^l + \mathcal{W}^l(\mathbf{h}_{\text{depth}}^l),
\end{equation}
where $\mathcal{W}^l$ is a learnable linear projection layer. This hierarchical feature fusion enables the RGB branch to maintain precise geometric structure while generating photorealistic video frames.
Both RGB and depth branches are optimized using the cross-entropy loss of tokens:
\begin{equation}
\begin{aligned}
\mathcal{L}_{\text{RGB}} = -\sum_{t=1}^T \mathbf{s}_t  \log p(\hat{\mathbf{s}}_t) , \quad
\mathcal{L}_{\text{Depth}} = -\sum_{t=1}^T \mathbf{z}_t  \log p(\hat{\mathbf{z}}_t) .
\end{aligned}
\end{equation}

\textbf{Implicit Material Understanding via Keypoint Dynamics Learning.}
Modeling physically plausible object deformations and motions in robot manipulation scenarios remains challenging for RGB-based world models, as material properties (e.g., rigidity, elasticity) cannot be effectively learned through RGB pixel fitting alone.
While physics engines provide accurate simulations, their computational expense and scene-specific constraints limit practical applicability.
To tackle this, we propose a keypoint-induced material learning approach, with the insight that physical material understanding can emerge from self-supervised tracking of contact-driven keypoint dynamics. 
For example, when a robot places an apple into a plastic bag, accurately capturing the motion of keypoints on the deforming bag implicitly captures the material properties. This method can be integrated naturally with video generation frameworks while maintaining strong generalization capabilities.

Specifically, for each video $\mathcal{V}$, we utilize SpatialTracker~\cite{xiao2024spatialtracker} to densely sample $N_0$ keypoints in the initial frame and track their temporal coordinate trajectories across $T$ frames, yielding $\mathcal{T}_{dense} = \{(\mathbf{p}_i^1, ..., \mathbf{p}_i^T)\}_{i=1}^{N_0}$, where the element $\mathbf{p}_i^t \in \mathbb{R}^2$ represents its coordinates in the tokenized feature map of frame $t$.
Rather than relying on costly segmentation masks to identify contact regions and guide keypoint sampling, we observe that the most informative keypoints are empirically characterized by large motion magnitudes. Thus, we adaptively select the top-$K$ most active keypoints based on their motion magnitudes $\mathcal{M}_i = \sum_{t=1}^{T-1} ||\mathbf{p}_i^{t+1} - \mathbf{p}_i^t||_2$, $\forall i \in {1,...,N_0}$, producing the sampled trajectory set $\mathcal{T}_{sample} = \{(\mathbf{p}_i^1, ..., \mathbf{p}_i^T)\}_{i=1}^{K}$.

To enhance the keypoint dynamic learning, we enforce temporal consistency between the visual tokens of sampled keypoints by aligning all frames to the initial frame ($t=1$) through the following loss:
\begin{equation}
\mathcal{L}_{\text{Keypoint}} = \frac{1}{(T-1)K}\sum_{i=1}^K\sum_{t=2}^T \|\hat{\mathbf{s}}_t(\mathbf{p}_i^t) - \hat{\mathbf{s}}_1(\mathbf{p}_i^1) \|_2^2,
\end{equation}
where $\hat{\mathbf{s}}_t(\mathbf{p}_i^t) \in \mathbb{R}^D$ denotes the $i$-th keypoint-located predicted token at frame $t$.

Furthermore, we observe that these dynamically active keypoint regions often exhibit higher token errors due to their complex motion patterns. 
To address this, we propose a keypoint-guided attention mechanism that adaptively enhances token learning in regions intersected by keypoint trajectories.
Specifically, we compute a spatiotemporal attention map $\mathbf{A} \in \mathbb{R}^{T \times H' \times W'}$, with each element defined as:
\begin{equation}
\mathbf{A}_{t,x,y} = \begin{cases} 
\gamma & \text{if } (t,x,y) \in \mathcal{T}_{sample}, \\
1 & \text{otherwise},
\end{cases}
\end{equation}
where $\gamma$ is a hyperparameter controlling the importance weight.
The attention-augmented training objective is formulated as:
\begin{equation}
\begin{aligned}
\mathcal{L}_{\text{Attention}} = -\sum_{t=1}^T \mathbf{A}_t \odot \mathbf{s}_t  \log p(\hat{\mathbf{s}}_t).
\end{aligned}
\end{equation}
\textbf{Physics-informed Joint Training Objectives.}
By integrating the above designs, we train a unified physics-aware world model through multi-task learning, with the final objective formulated as:
\begin{equation}
\begin{aligned}
\mathcal{L} = \mathcal{L}_{\text{RGB}} + \lambda_1 \mathcal{L}_{\text{Depth}} + \lambda_2 
 \mathcal{L}_{\text{Keypoint}} + \lambda_3 \mathcal{L}_{\text{Attention}},
\end{aligned}
\end{equation}
where $\lambda_1, \lambda_2, \lambda_3 \in \mathbb{R}^+$ are are tunable coefficients balancing the loss terms.

\section{Experiments}
In this section, we begin by detailing our experimental protocol (Section 4.1), including the dataset statistics, baseline information, and the implementation of our model. We then evaluate our model from three aspects: video quality evaluation (Section 4.2), robot policy learning with synthetic data (Section 4.3), and robotic policy evaluation (Section 4.4).
\subsection{Experimental Settings}
\textbf{Dataset Statistics.}
In our experiment, we use 50,000 video clips extracted from the AgiBotWorld-Beta dataset~\cite{contributors2025agibotworld}, covering 147 tasks and 72 skills. We concatenate the end position, end orientation, and effector position of the embodiment as the action sequence. 

\textbf{Baselines.}
We compare our model with four advanced baselines, including both embodied world models ({IRASim}~\cite{zhu2024irasim} and {iVideoGPT}~\cite{wu2024ivideogpt}) and general world models ({Genie}~\cite{bruce2024genie} and {CogVideoX}~\cite{yang2024cogvideox}). Due to unavailable training codes in some recent works~\cite{team2025aether,zhen2025tesseract}, these methods are excluded from direct comparison. Details of baselines are presented in the appendix.





\textbf{Implementation Details.}
We preprocess videos by extracting 16-frame clips sampled at 2Hz, yielding approximately 6.5 million training clips. 
The model is trained for 5 epochs using the following hyperparameters: $\lambda_1=1$, $\lambda_2=0.01$, $\lambda_3=1$, and $\gamma=5$. Training completes in approximately 24 hours on a cluster of 32 NVIDIA A800-SXM4-80GB GPUs. During inference, we use the first frame as a conditional input to autoregressively predict the subsequent 15 frames.

\subsection{Video Quality Evaluation}
\begin{figure}[t]
    \centering
    \includegraphics[width=\linewidth]{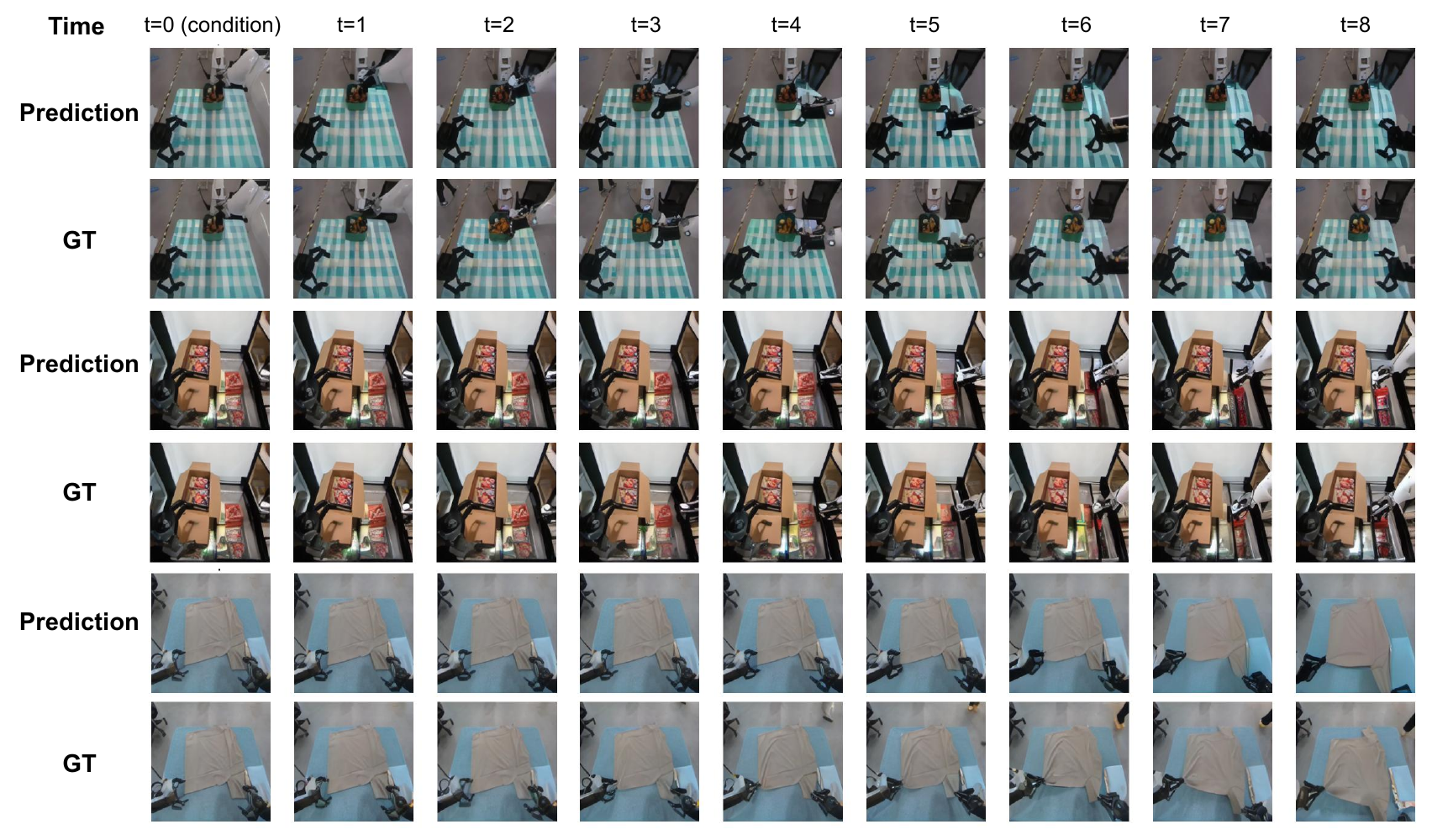}
    \caption{Qualitative results visualization of our model (only the subsequent 8 frames are shown). More results can be found in the appendix.}
    \label{fig:result}
\end{figure}

We evaluate video generation quality through three key dimensions: appearance fidelity, geometric consistency and action controllability. The details of the six used metrics are as follows:
\begin{itemize}[leftmargin=*]
    \item \textbf{PSNR}: It measures pixel-level reconstruction accuracy between generated and ground-truth frames.
    \item \textbf{LPIPS}: It assesses perceptual quality using visual feature similarity.
    \item \textbf{AbsRel}: It computes relative depth estimation errors.
    \item \textbf{$\mathbf{\delta_1/\delta_2}$}: They evaluate depth prediction accuracy at different precision levels.
    \item \textbf{$\Delta$PSNR}: It quantifies output sensitivity to action condition, with higher values indicating better action control ability.
    \end{itemize}
We present some generation results in Figure~\ref{fig:result}, where we predict future frames conditioned on an initial frame and robot action commands (we visualize 8 frames while the model supports long-horizon rollouts). The visualizations demonstrate that our model effectively simulates realistic robot manipulation scenarios, with generated sequences showing strong similarity to ground truth observations. Notably, our approach successfully handles deformable object interactions, as evidenced by the cloth-dragging sequence where the generated deformations accurately follow physical laws and capture material properties.

As shown in Table~\ref{tab:main_result}, we conduct comprehensive comparisons with four advanced baselines: two embodied world models (IRASim and iVideoGPT) and two general world models (Genie and CogVideoX). Our model consistently outperforms all baselines across six evaluation metrics, demonstrating its superior capability in video prediction for robotic scenarios.
Detailed analysis reveals that while CogVideoX can generate high-quality videos, its inability to follow action commands leads to substantial deviations in future frames. The two embodied world models are not good at motion learning when conducting long-term generation, thus receiving poor metrics. 
Our model's novel integration of keypoint dynamics learning effectively addresses these limitations, simultaneously achieving high-fidelity visual generation and superior action controllability.

\begin{figure}[t]
    \centering
    \includegraphics[width=0.95\linewidth]{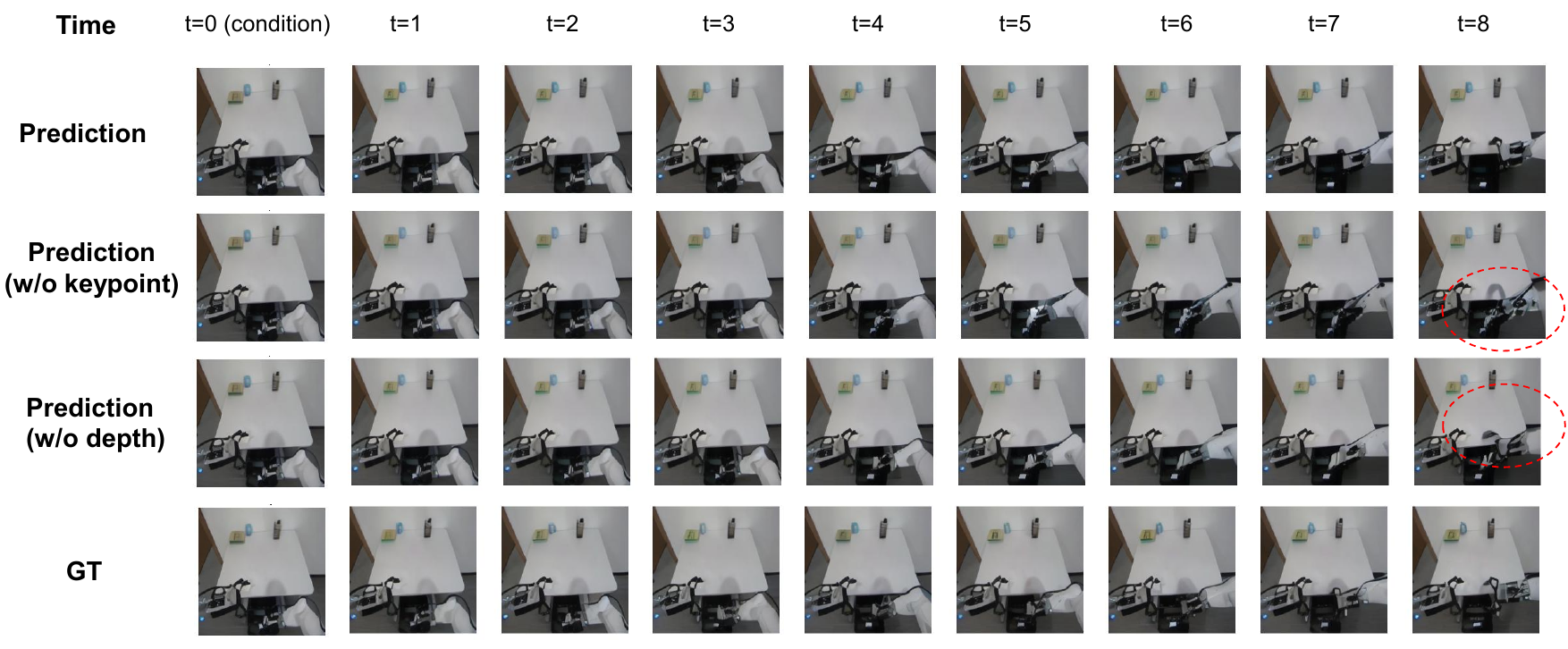}
    \caption{Effect of the physics knowledge learning. Omission of temporal depth learning leads to geometric distortions in moving objects, while the absence of key-point dynamics learning results in unreal motion patterns.}
    \label{fig:effect}
\end{figure}

We further conduct ablation studies to demonstrate the complementary benefits of our two core components: temporal depth learning and keypoint dynamics learning. The results are shown in Table~\ref{tab:ablation}. 
The quantitative results reveal that both components contribute significantly to overall performance; removing either one leads to measurable degradation across different metrics. 
The depth learning primarily preserves geometric consistency of moving objects, and the keypoint learning proves essential for maintaining both visual fidelity and action controllability. 
We provide a case study in Figure~\ref{fig:effect}. It can be seen that the missing of temporal depth learning will lead to geometric distortions in moving objects, while the absence of key-point dynamics learning results in unreal motion patterns. These findings collectively validate the necessity of our key designs.
\begin{table}[t]
    \centering
    \caption{Quantitative comparison of our model and baselines.}
    \begin{tabular}{cccccccc} 
        \hline
        \multirow{2}{*}{Method} & \multicolumn{2}{c}{Appearance Fidelity} & \multicolumn{3}{c}{Geometric Consistency} & Action Controllability\\
        \cline{2-7} 
         & LPIPS (↓) & PSNR (↑)  & AbsRel (↓) & $\delta_1$ (↑) & $\delta_2$ (↑) & $\Delta$PSNR (↑)\\ 
        \hline
        IRASim & 0.6674 & 11.5698 & 0.6252 & 0.5013 & 0.7020 &  0.0269 \\
        iVideoGPT & 0.4963 & 16.1236 & 0.7586 & 0.3480 & 0.5795  & 0.1144  \\
        Genie & 0.1683 & 19.7571 & 0.4425 & 0.5435 & 0.7736 &  1.9871 \\
        CogVideoX & 0.2180 & 17.5222 & 0.5243 & 0.6046 & 0.7599 & --- \\
        \textbf{RoboScape} & \textbf{0.1259} & \textbf{21.8533} & \textbf{0.3600} & \textbf{0.6214} & \textbf{0.8307} & \textbf{3.3435} \\
        \hline
    \end{tabular}
    \label{tab:main_result}
\end{table}

\begin{table}[t]
    \centering
    \caption{Ablation study of our key designs of physics prior injection.}
  \begin{tabularx}{\textwidth}{ccccccc} 
        \hline
        Method & LPIPS (↓) & PSNR (↑)  & AbsRel (↓) & $\delta_1$ (↑) & $\delta_2$ (↑) & $\Delta$PSNR (↑)\\ 
        \hline
        whole model & 0.1259 & 21.8533 & 0.3600 & 0.6214 & 0.8307 & 3.3435 \\
        w/o depth & 0.1249 & 21.9465 & 0.3921 & 0.5788 & 0.8277 & 3.4863 \\
        w/o keypoint & 0.1264 & 21.7087 & 0.3417 & 0.6497 & 0.8673 & 2.9462 \\
        w/o depth \& keypoint & 0.1299 & 21.4873
 & 0.3565 & 0.6248 &  0.8129 & 1.9871\\
        \hline
    \end{tabularx}
    \label{tab:ablation}
\end{table}

\subsection{Robotic Policy Learning with Synthetic Data}
We validate our world model's utility by generating synthetic robotic video data for downstream policy learning based on Diffusion Policy (DP)~\cite{chi2023diffusion} and $\pi$0~\cite{black2024pi_0}. Through controlled experiments with progressively adding synthetic data, we systematically measure the impact of generated data on policy learning performance. The results are shown in Table~\ref{tab:policy_learning}.
\begin{table}[h]
    \centering
    \small
    \caption{Results of policy learning with DP on Robomimic task and $\pi$0 on LIBERO tasks.}
    \resizebox{\linewidth}{!}{
    \begin{tabularx}{\textwidth}{ccccccccc} 
        \cline{1-2} \cline{4-9}
        \multicolumn{2}{c}{DP on Robomimic tasks} & & \multicolumn{6}{c}{$\pi$0 on LIBERO tasks} \\
        \cline{1-2} \cline{4-9}
        \# Synthetic Data & Success Rate & & \#Synthetic Data & Spatial &Object &Goal &10&  Average \\
        \cline{1-2} \cline{4-9} 
        50& 40\% & &200   & 77.6\%   & 81.8\%   & 71.0\%   & 36.0\%   & 66.6\%   \\
        100& 77\% & &400   & 79.4\%   & 85.2\%   & 74.6\%   & 46.2\%   & 71.4\%   \\
        150& 84\% & &600   & 81.6\%   & 86.0\%   & 78.0\%   & 51.8\%   & 74.4\%   \\
        200& 91\% & &800   & 84.6\%   & 89.0\%   & 82.8\%   & 60.0\%   & 79.1\%   \\
        \cline{1-2} \cline{4-9}
        Real (200) & 92\% & & Real (200) & 77.2\%   & 79.8\%   & 68.8\%   & 34.8\%   & 65.2\%   \\
        \cline{1-2} \cline{4-9}
    \end{tabularx}
    }
    \label{tab:policy_learning}
\end{table}

In the experiments on the Robomimic Lift task~\cite{mandlekar2021matters}, DP trained for 10k steps with only generated data achieved nearly the same performance as DP trained with real data. 
Notably, the policy success rate exhibited consistent improvement with increasing synthetic training data, highlighting the effectiveness of our model.
We further validated our approach using the 
$\pi_0$~\cite{black2024pi_0} model on the challenging LIBERO~\cite{liu2023libero} task suite. 
These tasks present three key challenges beyond the Robomimic Lift environment: (1) complex multi-object manipulation requirements, (2) cluttered scene configurations, and (3) extended action sequence horizons. Therefore, we employ a small amount of real data (200 trajectories) as a training warm-up.
Remarkably, when training $\pi_0$ policies with increasing generated data, the model performance achieves gradual improvement.
These results demonstrate our model's capability to generate physically plausible trajectories even for demanding, long-horizon manipulation scenarios.
\begin{figure}[t] 
	\centering  
	\subfigure[IRASim]{
        \includegraphics[width=0.32\linewidth]{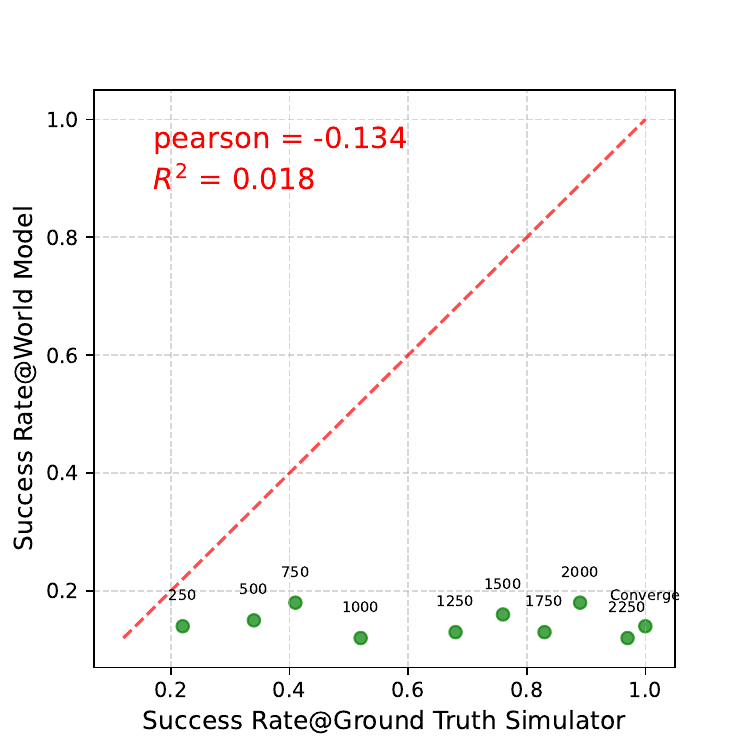}}
    \subfigure[iVideoGPT]{
	\includegraphics[width=0.32\linewidth]{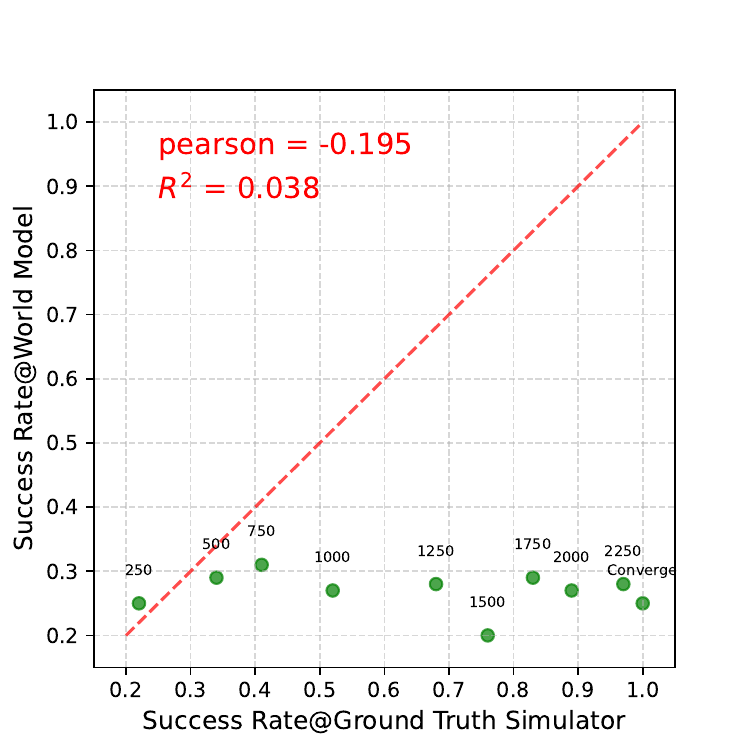}}
    \subfigure[RoboScape]{
	\includegraphics[width=0.32\linewidth]{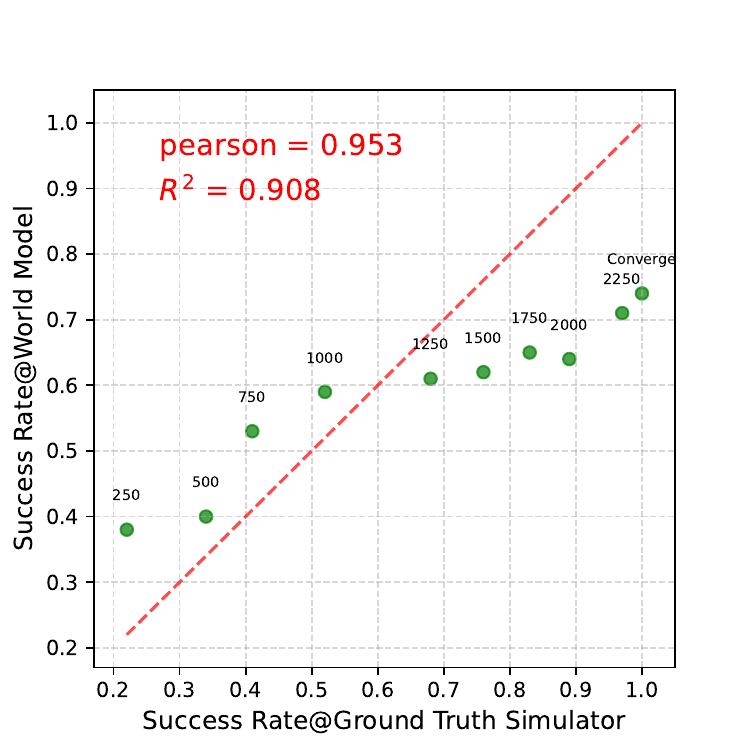}}
    \caption{Correlation between the success rate of different world models and the ground-truth simulator. Each point represents a policy, and the trained epochs are shown above the point.}
	\label{fig:policy_evaluation}
\end{figure}
\subsection{Robotic Policy Evaluation}
In this section, we investigate whether our world model can act as a policy evaluator for different robotic policies. In policy evaluation, the world model acts as an environment that receives policy-generated action sequences and predicts subsequent observations in a rollout manner. Policy quality is then assessed by checking success rates in the predicted videos.
Here we compare IRASim, iVideoGPT, and our model as the policy evaluator and use Diffusion Policy~\cite{chi2023diffusion} as the policy model. 
Specifically, we train the policy on the Robomimic Lift task~\cite{mandlekar2021matters} using 200 trajectories and save the policy every 250 epochs until it is fully converged. Then we post-train the world model and evaluate the policy in both the ground-truth simulator and the world model by 100 runs. 
The success signal of each run can be directly given by the simulator, while it requires manual judgment when the policy interacts with the world model.
Afterwards, we calculate the Pearson correlation and $R^2$ between different world models and the ground-truth simulator. The results in Figure~\ref{fig:policy_evaluation} show that the Pearson correlation of our model is \textbf{0.953}, while the correlation of other models is rather low, indicating that our world model can be utilized as a better policy evaluator.
\section{Related Work}
\subsection{World Model}
World models learn representations of environmental states through neural networks, enabling the prediction of future states based on current observations and actions~\citep{ha2018world}. 
Recent advances in world models primarily leverage video generation techniques, with applications spanning three key domains including autonomous driving~\citep{wang2024drivedreamer,zheng2024genad,che2024gamegen,ni2025maskgwm,wang2024driving,gao2024vista,russell2025gaia}, embodied intelligence~\citep{wang2025hma,zhu2024irasim,agarwal2025cosmos}, and gaming~\citep{yu2025gamefactory,he2025pre,xiao2025worldmem,che2024gamegen}. The dominant modeling approaches fall into two main categories: diffusion models and autoregressive models. Diffusion models, such as DiT~\cite{peebles2023scalable}, generate sequences through a gradual denoising process and are well-suited for producing diverse and short-term consistent visual content. 
Autoregressive models, such as Genie\cite{bruce2024genie}, reconstruct sequences via masking mechanisms and demonstrate superior efficiency and controllability. 
Our work utilizes masked autoregressive models with physical information injection, aiming to build efficient and interactive world models.
\subsection{Physics-aware Generative Model}
Recent advances in video generation have increasingly focused on improving the modeling of physical properties~\cite{lin2025exploring}. Current methods in this area can be roughly divided into explicit and implicit physical modeling. Explicit methods incorporate physical information by learning explicit textures and material representations~\cite{liu2024physics3d,zhen2025tesseract}. In contrast, implicit methods mainly embed physical knowledge into models via training loss terms~\cite{chen2025towards}, or by using generative models to jointly generate RGB videos and other physical representations~\citep{team2025aether,liang2025UniFuture}. 
These approaches aim to enhance physical understanding through data-driven approaches rather than predefined physical rules.
Currently, there's much room for existing embodied world models to enhance the integration of physical knowledge into video generation. To advance this field, we introduce a physics-informed embodied world model that jointly learns RGB video generation, temporal depth prediction, and keypoint dynamics within a unified framework, achieving both high visual fidelity and physical plausibility.

\section{Conclusion and Future Work}
In this work, we propose RoboScape, a physics-informed embodied world model that efficiently integrates physical knowledge into video generation through a physics-inspired multi-task joint training framework, eliminating the need for cascaded external models such as physics engines. 
By incorporating temporal depth prediction, our model learns the 3D geometric structure of scenes, while dynamic keypoint learning enables implicit modeling of object deformation and motion patterns. Extensive evaluations demonstrate that our approach outperforms baseline methods in video generation quality, synthetic data utility for downstream robotic manipulation policy training, and effectiveness as a policy evaluator. 
In the future, we plan to combine the generative world model with real-world robots to test performance further.

\clearpage
\bibliographystyle{unsrt} 
\bibliography{ref}

\begin{thebibliography}{10}

\bibitem{achiam2023gpt}
Josh Achiam, Steven Adler, Sandhini Agarwal, Lama Ahmad, Ilge Akkaya, Florencia~Leoni Aleman, Diogo Almeida, Janko Altenschmidt, Sam Altman, Shyamal Anadkat, et~al.
\newblock Gpt-4 technical report.
\newblock {\em arXiv preprint arXiv:2303.08774}, 2023.

\bibitem{team2023gemini}
Gemini Team, Rohan Anil, Sebastian Borgeaud, Jean-Baptiste Alayrac, Jiahui Yu, Radu Soricut, Johan Schalkwyk, Andrew~M Dai, Anja Hauth, Katie Millican, et~al.
\newblock Gemini: a family of highly capable multimodal models.
\newblock {\em arXiv preprint arXiv:2312.11805}, 2023.

\bibitem{walke2023bridgedata}
Homer~Rich Walke, Kevin Black, Tony~Z Zhao, Quan Vuong, Chongyi Zheng, Philippe Hansen-Estruch, Andre~Wang He, Vivek Myers, Moo~Jin Kim, Max Du, et~al.
\newblock Bridgedata v2: A dataset for robot learning at scale.
\newblock In {\em Conference on Robot Learning}, pages 1723--1736. PMLR, 2023.

\bibitem{o2024open}
Abby O’Neill, Abdul Rehman, Abhiram Maddukuri, Abhishek Gupta, Abhishek Padalkar, Abraham Lee, Acorn Pooley, Agrim Gupta, Ajay Mandlekar, Ajinkya Jain, et~al.
\newblock Open x-embodiment: Robotic learning datasets and rt-x models: Open x-embodiment collaboration 0.
\newblock In {\em 2024 IEEE International Conference on Robotics and Automation (ICRA)}, pages 6892--6903. IEEE, 2024.

\bibitem{khazatsky2024droid}
Alexander Khazatsky, Karl Pertsch, Suraj Nair, Ashwin Balakrishna, Sudeep Dasari, Siddharth Karamcheti, Soroush Nasiriany, Mohan~Kumar Srirama, Lawrence~Yunliang Chen, Kirsty Ellis, et~al.
\newblock Droid: A large-scale in-the-wild robot manipulation dataset.
\newblock {\em arXiv preprint arXiv:2403.12945}, 2024.

\bibitem{bu2025agibot}
Qingwen Bu, Jisong Cai, Li~Chen, Xiuqi Cui, Yan Ding, Siyuan Feng, Shenyuan Gao, Xindong He, Xu~Huang, Shu Jiang, et~al.
\newblock Agibot world colosseo: A large-scale manipulation platform for scalable and intelligent embodied systems.
\newblock {\em arXiv preprint arXiv:2503.06669}, 2025.

\bibitem{ha2018recurrent}
David Ha and J{\"u}rgen Schmidhuber.
\newblock Recurrent world models facilitate policy evolution.
\newblock {\em Advances in neural information processing systems}, 31, 2018.

\bibitem{yang2023learning}
Mengjiao Yang, Yilun Du, Kamyar Ghasemipour, Jonathan Tompson, Dale Schuurmans, and Pieter Abbeel.
\newblock Learning interactive real-world simulators.
\newblock {\em arXiv preprint arXiv:2310.06114}, 1(2):6, 2023.

\bibitem{brooks2024video}
Tim Brooks, Bill Peebles, Connor Holmes, Will DePue, Yufei Guo, Li~Jing, David Schnurr, Joe Taylor, Troy Luhman, Eric Luhman, et~al.
\newblock Video generation models as world simulators.
\newblock {\em OpenAI Blog}, 1:8, 2024.

\bibitem{wang2025learning}
Lirui Wang, Kevin Zhao, Chaoqi Liu, and Xinlei Chen.
\newblock Learning re al-world action-video dynamics with heterogeneous masked autoregression.
\newblock {\em arXiv preprint arXiv:2502.04296}, 2025.

\bibitem{wu2024ivideogpt}
Jialong Wu, Shaofeng Yin, Ningya Feng, Xu~He, Dong Li, Jianye Hao, and Mingsheng Long.
\newblock ivideogpt: Interactive videogpts are scalable world models.
\newblock {\em Advances in Neural Information Processing Systems}, 37:68082--68119, 2024.

\bibitem{zhu2024irasim}
Fangqi Zhu, Hongtao Wu, Song Guo, Yuxiao Liu, Chilam Cheang, and Tao Kong.
\newblock Irasim: Learning interactive real-robot action simulators.
\newblock {\em arXiv preprint arXiv:2406.14540}, 2024.

\bibitem{agarwal2025cosmos}
Niket Agarwal, Arslan Ali, Maciej Bala, Yogesh Balaji, Erik Barker, Tiffany Cai, Prithvijit Chattopadhyay, Yongxin Chen, Yin Cui, Yifan Ding, et~al.
\newblock Cosmos world foundation model platform for physical ai.
\newblock {\em arXiv preprint arXiv:2501.03575}, 2025.

\bibitem{kang2024far}
Bingyi Kang, Yang Yue, Rui Lu, Zhijie Lin, Yang Zhao, Kaixin Wang, Gao Huang, and Jiashi Feng.
\newblock How far is video generation from world model: A physical law perspective.
\newblock {\em arXiv preprint arXiv:2411.02385}, 2024.

\bibitem{lin2025exploring}
Minghui Lin, Xiang Wang, Yishan Wang, Shu Wang, Fengqi Dai, Pengxiang Ding, Cunxiang Wang, Zhengrong Zuo, Nong Sang, Siteng Huang, et~al.
\newblock Exploring the evolution of physics cognition in video generation: A survey.
\newblock {\em arXiv preprint arXiv:2503.21765}, 2025.

\bibitem{guo2025t2vphysbench}
Xuyang Guo, Jiayan Huo, Zhenmei Shi, Zhao Song, Jiahao Zhang, and Jiale Zhao.
\newblock T2vphysbench: A first-principles benchmark for physical consistency in text-to-video generation.
\newblock {\em arXiv preprint arXiv:2505.00337}, 2025.

\bibitem{meng2024towards}
Fanqing Meng, Jiaqi Liao, Xinyu Tan, Wenqi Shao, Quanfeng Lu, Kaipeng Zhang, Yu~Cheng, Dianqi Li, Yu~Qiao, and Ping Luo.
\newblock Towards world simulator: Crafting physical commonsense-based benchmark for video generation.
\newblock {\em arXiv preprint arXiv:2410.05363}, 2024.

\bibitem{chen2025towards}
Yunuo Chen, Junli Cao, Anil Kag, Vidit Goel, Sergei Korolev, Chenfanfu Jiang, Sergey Tulyakov, and Jian Ren.
\newblock Towards physical understanding in video generation: A 3d point regularization approach.
\newblock {\em arXiv preprint arXiv:2502.03639}, 2025.

\bibitem{rai2024enhancing}
Gaurav Rai and Ojaswa Sharma.
\newblock Enhancing sketch animation: Text-to-video diffusion models with temporal consistency and rigidity constraints.
\newblock {\em arXiv preprint arXiv:2411.19381}, 2024.

\bibitem{luiten2024dynamic}
Jonathon Luiten, Georgios Kopanas, Bastian Leibe, and Deva Ramanan.
\newblock Dynamic 3d gaussians: Tracking by persistent dynamic view synthesis.
\newblock In {\em 2024 International Conference on 3D Vision (3DV)}, pages 800--809. IEEE, 2024.

\bibitem{xie2025physanimator}
Tianyi Xie, Yiwei Zhao, Ying Jiang, and Chenfanfu Jiang.
\newblock Physanimator: Physics-guided generative cartoon animation.
\newblock {\em arXiv preprint arXiv:2501.16550}, 2025.

\bibitem{lv2024gpt4motion}
Jiaxi Lv, Yi~Huang, Mingfu Yan, Jiancheng Huang, Jianzhuang Liu, Yifan Liu, Yafei Wen, Xiaoxin Chen, and Shifeng Chen.
\newblock Gpt4motion: Scripting physical motions in text-to-video generation via blender-oriented gpt planning.
\newblock In {\em Proceedings of the IEEE/CVF conference on computer vision and pattern recognition}, pages 1430--1440, 2024.

\bibitem{liu2024physgen}
Shaowei Liu, Zhongzheng Ren, Saurabh Gupta, and Shenlong Wang.
\newblock Physgen: Rigid-body physics-grounded image-to-video generation.
\newblock In {\em European Conference on Computer Vision}, pages 360--378. Springer, 2024.

\bibitem{tan2024physmotion}
Xiyang Tan, Ying Jiang, Xuan Li, Zeshun Zong, Tianyi Xie, Yin Yang, and Chenfanfu Jiang.
\newblock Physmotion: Physics-grounded dynamics from a single image.
\newblock {\em arXiv preprint arXiv:2411.17189}, 2024.

\bibitem{zhang2024physdreamer}
Tianyuan Zhang, Hong-Xing Yu, Rundi Wu, Brandon~Y Feng, Changxi Zheng, Noah Snavely, Jiajun Wu, and William~T Freeman.
\newblock Physdreamer: Physics-based interaction with 3d objects via video generation.
\newblock In {\em European Conference on Computer Vision}, pages 388--406. Springer, 2024.

\bibitem{liu2024physics3d}
Fangfu Liu, Hanyang Wang, Shunyu Yao, Shengjun Zhang, Jie Zhou, and Yueqi Duan.
\newblock Physics3d: Learning physical properties of 3d gaussians via video diffusion.
\newblock {\em arXiv preprint arXiv:2406.04338}, 2024.

\bibitem{team2025aether}
Aether Team, Haoyi Zhu, Yifan Wang, Jianjun Zhou, Wenzheng Chang, Yang Zhou, Zizun Li, Junyi Chen, Chunhua Shen, Jiangmiao Pang, et~al.
\newblock Aether: Geometric-aware unified world modeling.
\newblock {\em arXiv preprint arXiv:2503.18945}, 2025.

\bibitem{zhen2025tesseract}
Haoyu Zhen, Qiao Sun, Hongxin Zhang, Junyan Li, Siyuan Zhou, Yilun Du, and Chuang Gan.
\newblock Tesseract: Learning 4d embodied world models.
\newblock {\em arXiv preprint arXiv:2504.20995}, 2025.

\bibitem{chi2023diffusion}
Cheng Chi, Zhenjia Xu, Siyuan Feng, Eric Cousineau, Yilun Du, Benjamin Burchfiel, Russ Tedrake, and Shuran Song.
\newblock Diffusion policy: Visuomotor policy learning via action diffusion.
\newblock {\em The International Journal of Robotics Research}, page 02783649241273668, 2023.

\bibitem{black2024pi_0}
Kevin Black, Noah Brown, Danny Driess, Adnan Esmail, Michael Equi, Chelsea Finn, Niccolo Fusai, Lachy Groom, Karol Hausman, Brian Ichter, et~al.
\newblock $\pi_0$: A vision-language-action flow model for general robot control.
\newblock {\em arXiv preprint arXiv:2410.24164}, 2024.

\bibitem{chen2025video}
Sili Chen, Hengkai Guo, Shengnan Zhu, Feihu Zhang, Zilong Huang, Jiashi Feng, and Bingyi Kang.
\newblock Video depth anything: Consistent depth estimation for super-long videos.
\newblock {\em arXiv preprint arXiv:2501.12375}, 2025.

\bibitem{xiao2024spatialtracker}
Yuxi Xiao, Qianqian Wang, Shangzhan Zhang, Nan Xue, Sida Peng, Yujun Shen, and Xiaowei Zhou.
\newblock Spatialtracker: Tracking any 2d pixels in 3d space.
\newblock In {\em Proceedings of the IEEE/CVF Conference on Computer Vision and Pattern Recognition}, pages 20406--20417, 2024.

\bibitem{soucek2024transnet}
Tom{\'a}s Soucek and Jakub Lokoc.
\newblock Transnet v2: An effective deep network architecture for fast shot transition detection.
\newblock In {\em Proceedings of the 32nd ACM International Conference on Multimedia}, pages 11218--11221, 2024.

\bibitem{chen2024internvl}
Zhe Chen, Jiannan Wu, Wenhai Wang, Weijie Su, Guo Chen, Sen Xing, Muyan Zhong, Qinglong Zhang, Xizhou Zhu, Lewei Lu, et~al.
\newblock Internvl: Scaling up vision foundation models and aligning for generic visual-linguistic tasks.
\newblock In {\em Proceedings of the IEEE/CVF conference on computer vision and pattern recognition}, pages 24185--24198, 2024.

\bibitem{dosovitskiy2015flownet}
Alexey Dosovitskiy, Philipp Fischer, Eddy Ilg, Philip Hausser, Caner Hazirbas, Vladimir Golkov, Patrick Van Der~Smagt, Daniel Cremers, and Thomas Brox.
\newblock Flownet: Learning optical flow with convolutional networks.
\newblock In {\em Proceedings of the IEEE international conference on computer vision}, pages 2758--2766, 2015.

\bibitem{wang2021survey}
Xin Wang, Yudong Chen, and Wenwu Zhu.
\newblock A survey on curriculum learning.
\newblock {\em IEEE transactions on pattern analysis and machine intelligence}, 44(9):4555--4576, 2021.

\bibitem{yu2023language}
Lijun Yu, Jos{\'e} Lezama, Nitesh~B Gundavarapu, Luca Versari, Kihyuk Sohn, David Minnen, Yong Cheng, Vighnesh Birodkar, Agrim Gupta, Xiuye Gu, et~al.
\newblock Language model beats diffusion--tokenizer is key to visual generation.
\newblock {\em arXiv preprint arXiv:2310.05737}, 2023.

\bibitem{contributors2025agibotworld}
AgiBot-World-Contributors, Qingwen Bu, Jisong Cai, Li~Chen, Xiuqi Cui, Yan Ding, Siyuan Feng, Shenyuan Gao, Xindong He, Xuan Hu, Xu~Huang, Shu Jiang, Yuxin Jiang, Cheng Jing, Hongyang Li, Jialu Li, Chiming Liu, Yi~Liu, Yuxiang Lu, Jianlan Luo, Ping Luo, Yao Mu, Yuehan Niu, Yixuan Pan, Jiangmiao Pang, Yu~Qiao, Guanghui Ren, Cheng Ruan, Jiaqi Shan, Yongjian Shen, Chengshi Shi, Mingkang Shi, Modi Shi, Chonghao Sima, Jianheng Song, Huijie Wang, Wenhao Wang, Dafeng Wei, Chengen Xie, Guo Xu, Junchi Yan, Cunbiao Yang, Lei Yang, Shukai Yang, Maoqing Yao, Jia Zeng, Chi Zhang, Qinglin Zhang, Bin Zhao, Chengyue Zhao, Jiaqi Zhao, and Jianchao Zhu.
\newblock Agibot world colosseo: A large-scale manipulation platform for scalable and intelligent embodied systems.
\newblock {\em arXiv preprint arXiv:2503.06669}, 2025.

\bibitem{bruce2024genie}
Jake Bruce, Michael~D Dennis, Ashley Edwards, Jack Parker-Holder, Yuge Shi, Edward Hughes, Matthew Lai, Aditi Mavalankar, Richie Steigerwald, Chris Apps, et~al.
\newblock Genie: Generative interactive environments.
\newblock In {\em Forty-first International Conference on Machine Learning}, 2024.

\bibitem{yang2024cogvideox}
Zhuoyi Yang, Jiayan Teng, Wendi Zheng, Ming Ding, Shiyu Huang, Jiazheng Xu, Yuanming Yang, Wenyi Hong, Xiaohan Zhang, Guanyu Feng, et~al.
\newblock Cogvideox: Text-to-video diffusion models with an expert transformer.
\newblock {\em arXiv preprint arXiv:2408.06072}, 2024.

\bibitem{mandlekar2021matters}
Ajay Mandlekar, Danfei Xu, Josiah Wong, Soroush Nasiriany, Chen Wang, Rohun Kulkarni, Li~Fei-Fei, Silvio Savarese, Yuke Zhu, and Roberto Mart{\'\i}n-Mart{\'\i}n.
\newblock What matters in learning from offline human demonstrations for robot manipulation.
\newblock {\em arXiv preprint arXiv:2108.03298}, 2021.

\bibitem{liu2023libero}
Bo~Liu, Yifeng Zhu, Chongkai Gao, Yihao Feng, Qiang Liu, Yuke Zhu, and Peter Stone.
\newblock Libero: Benchmarking knowledge transfer for lifelong robot learning.
\newblock {\em Advances in Neural Information Processing Systems}, 36:44776--44791, 2023.

\bibitem{ha2018world}
David Ha and J{\"u}rgen Schmidhuber.
\newblock World models.
\newblock {\em arXiv preprint arXiv:1803.10122}, 2018.

\bibitem{wang2024drivedreamer}
Xiaofeng Wang, Zheng Zhu, Guan Huang, Xinze Chen, Jiagang Zhu, and Jiwen Lu.
\newblock Drivedreamer: Towards real-world-drive world models for autonomous driving.
\newblock In {\em European Conference on Computer Vision}, pages 55--72. Springer, 2024.

\bibitem{zheng2024genad}
Wenzhao Zheng, Ruiqi Song, Xianda Guo, Chenming Zhang, and Long Chen.
\newblock Genad: Generative end-to-end autonomous driving.
\newblock In {\em European Conference on Computer Vision}, pages 87--104. Springer, 2024.

\bibitem{che2024gamegen}
Haoxuan Che, Xuanhua He, Quande Liu, Cheng Jin, and Hao Chen.
\newblock Gamegen-x: Interactive open-world game video generation.
\newblock {\em arXiv preprint arXiv:2411.00769}, 2024.

\bibitem{ni2025maskgwm}
Jingcheng Ni, Yuxin Guo, Yichen Liu, Rui Chen, Lewei Lu, and Zehuan Wu.
\newblock Maskgwm: A generalizable driving world model with video mask reconstruction.
\newblock {\em arXiv preprint arXiv:2502.11663}, 2025.

\bibitem{wang2024driving}
Yuqi Wang, Jiawei He, Lue Fan, Hongxin Li, Yuntao Chen, and Zhaoxiang Zhang.
\newblock Driving into the future: Multiview visual forecasting and planning with world model for autonomous driving.
\newblock In {\em Proceedings of the IEEE/CVF Conference on Computer Vision and Pattern Recognition}, pages 14749--14759, 2024.

\bibitem{gao2024vista}
Shenyuan Gao, Jiazhi Yang, Li~Chen, Kashyap Chitta, Yihang Qiu, Andreas Geiger, Jun Zhang, and Hongyang Li.
\newblock Vista: A generalizable driving world model with high fidelity and versatile controllability.
\newblock {\em arXiv preprint arXiv:2405.17398}, 2024.

\bibitem{russell2025gaia}
Lloyd Russell, Anthony Hu, Lorenzo Bertoni, George Fedoseev, Jamie Shotton, Elahe Arani, and Gianluca Corrado.
\newblock Gaia-2: A controllable multi-view generative world model for autonomous driving.
\newblock {\em arXiv preprint arXiv:2503.20523}, 2025.

\bibitem{wang2025hma}
Lirui Wang, Kevin Zhao, Chaoqi Liu, and Xinlei Chen.
\newblock Learning robotic video dynamics with heterogeneous masked autoregression.
\newblock In {\em Arxiv}, 2025.

\bibitem{yu2025gamefactory}
Jiwen Yu, Yiran Qin, Xintao Wang, Pengfei Wan, Di~Zhang, and Xihui Liu.
\newblock Gamefactory: Creating new games with generative interactive videos.
\newblock {\em arXiv preprint arXiv:2501.08325}, 2025.

\bibitem{he2025pre}
Haoran He, Yang Zhang, Liang Lin, Zhongwen Xu, and Ling Pan.
\newblock Pre-trained video generative models as world simulators.
\newblock {\em arXiv preprint arXiv:2502.07825}, 2025.

\bibitem{xiao2025worldmem}
Zeqi Xiao, Yushi Lan, Yifan Zhou, Wenqi Ouyang, Shuai Yang, Yanhong Zeng, and Xingang Pan.
\newblock Worldmem: Long-term consistent world simulation with memory.
\newblock {\em arXiv preprint arXiv:2504.12369}, 2025.

\bibitem{peebles2023scalable}
William Peebles and Saining Xie.
\newblock Scalable diffusion models with transformers.
\newblock In {\em Proceedings of the IEEE/CVF international conference on computer vision}, pages 4195--4205, 2023.

\bibitem{liang2025UniFuture}
Dingkang Liang, Dingyuan Zhang, Xin Zhou, Sifan Tu, Tianrui Feng, Xiaofan Li, Yumeng Zhang, Mingyang Du, Xiao Tan, and Xiang Bai.
\newblock Seeing the future, perceiving the future: A unified driving world model for future generation and perception.
\newblock {\em arXiv preprint arXiv:2503.13587}, 2025.

\end{thebibliography}
\newpage
\appendix

\section{Broader Impacts}
Our physics-informed world model significantly advances robotic learning by generating high-fidelity synthetic data with inherent physical plausibility, reducing reliance on costly real-world data collection while improving simulation-to-reality transfer. This technology enables safer and more efficient training of assistive robots for healthcare, disaster response, and industrial applications, while its computational efficiency lowers barriers for broader research participation. By embedding physical constraints during generation, we enhance the reliability of robotic data generation, though future work should further address considerations in synthetic data diversity and establish governance frameworks for responsible deployment in safety-critical domains.


\section{Baseline Details}
We provide details of the compared baselines as follows:
\begin{itemize}[leftmargin=*]

\item \textbf{IRASim}: A DiT-based robotic video generation model, capable of generating videos conditioned on robot actions and trajectories.

\item \textbf{iVideoGPT}: An auto-regressive interactive world model that takes the current video frame observation and action as input to predict the next frame while simultaneously estimating the reward signal for robotic operations.

\item \textbf{Genie}: A foundation world model trained through unsupervised learning on massive video data. We implement it with a reproduced open-source repository~\footnote{https://github.com/1x-technologies/1xgpt}.

\item \textbf{CogVideoX}: An advanced DiT-based text-to-video generation framework, with superior performance in prompt-driven video generation. 
\end{itemize}
\section{Scaling Behavior of RoboScape}
\begin{figure}[t]
    \centering
    \includegraphics[width=\linewidth]{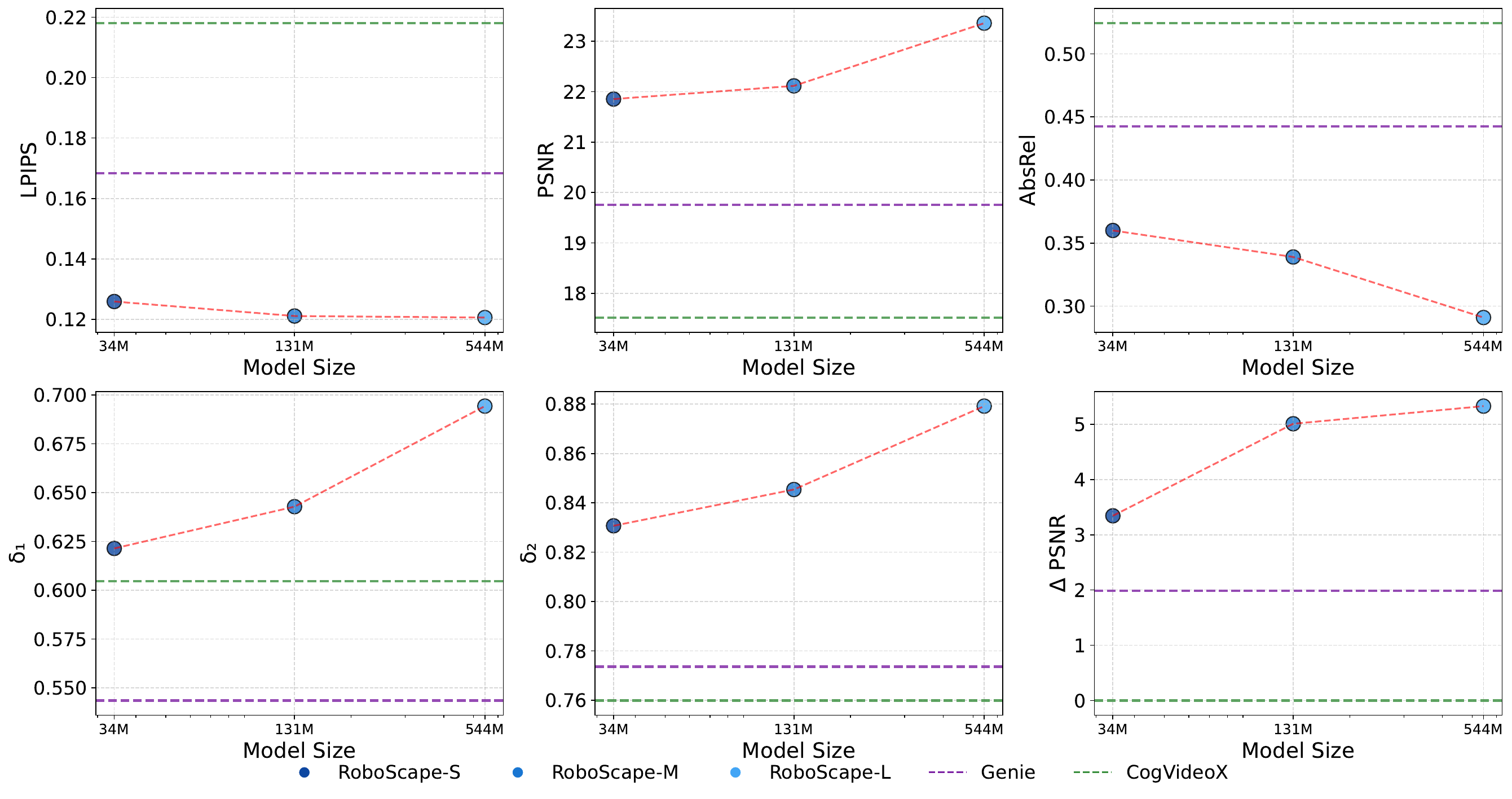}
    \caption{Model scaling law of RoboScape. }
    \label{fig:model_scale}
\end{figure}

\begin{figure}[t]
    \centering
    \includegraphics[width=\linewidth]{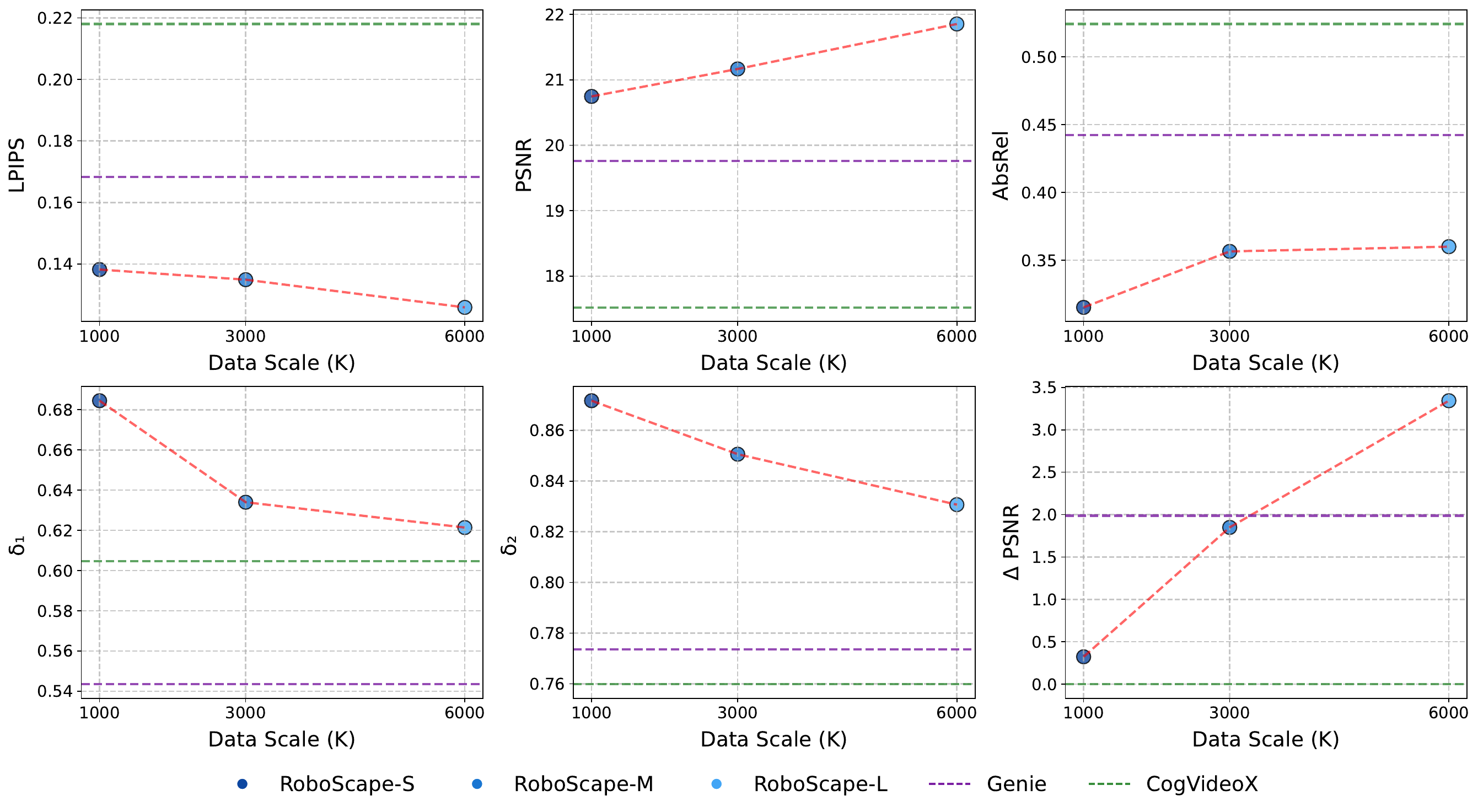}
    \caption{Data scaling law of RoboScape-S. }
    \label{fig:data_scale}
\end{figure}

We investigate the scaling behavior of RoboScape in terms of both model and data scales. As shown in Figure~\ref{fig:model_scale}, we evaluate three model variants—RoboScape-S (34M), RoboScape-M (131M), and RoboScape-L (544M)—and observe a clear scaling law: all six evaluation metrics improve significantly as model capacity increases. 

In addition, we study the impact of data scale by training RoboScape-S on 1,000K, 3,000K, and 6,000K clips (Figure~\ref{fig:data_scale}). While increasing data size consistently enhances visual quality and action controllability, geometric accuracy exhibits marginal improvement or even slight degradation. We find that this is because smaller datasets encourage overfitting to the final frame of conditional inputs, artificially inflating geometric metrics without generating meaningful temporal dynamics. 
Despite this, the overall trend confirms that more training data leads to better model performance.  

These findings highlight the importance of both model and data scaling in advancing robotic video generation, with larger models and datasets yielding better results.

\section{More Visualization Results}
\subsection{Video Generation Results}
\begin{figure}[t]
    \centering
    \includegraphics[width=\linewidth]{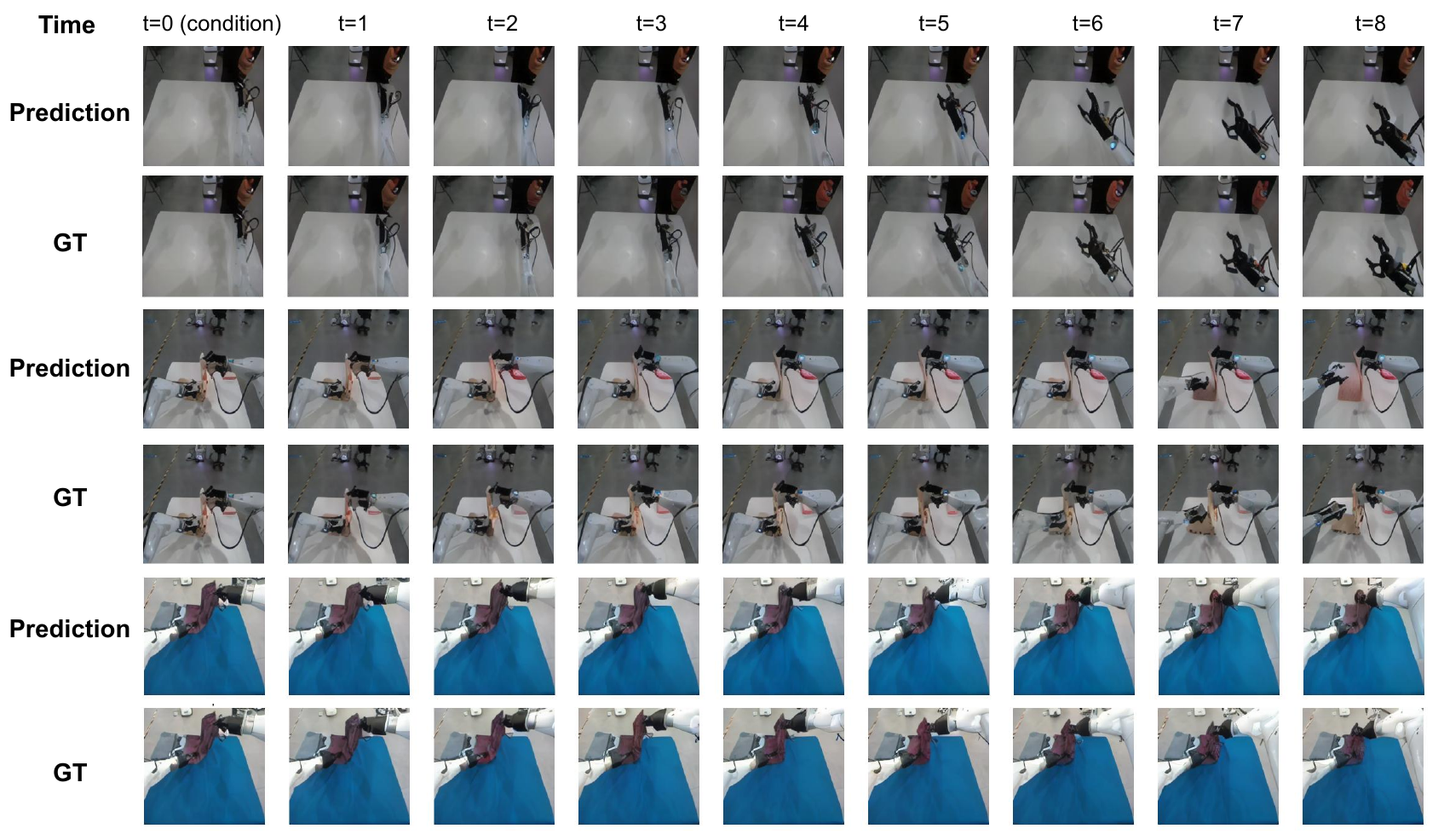}
    \caption{Supplemented visualization results from our model (only the subsequent 8 frames are shown). }
    \label{fig:result_supp}
\end{figure}

We provide more visualization results of generated videos using our model, as illustrated in Figure~\ref{fig:result_supp}.

\subsection{Robotic Policy Learning}

\begin{figure}[t]
    \centering
    \includegraphics[width=\linewidth]{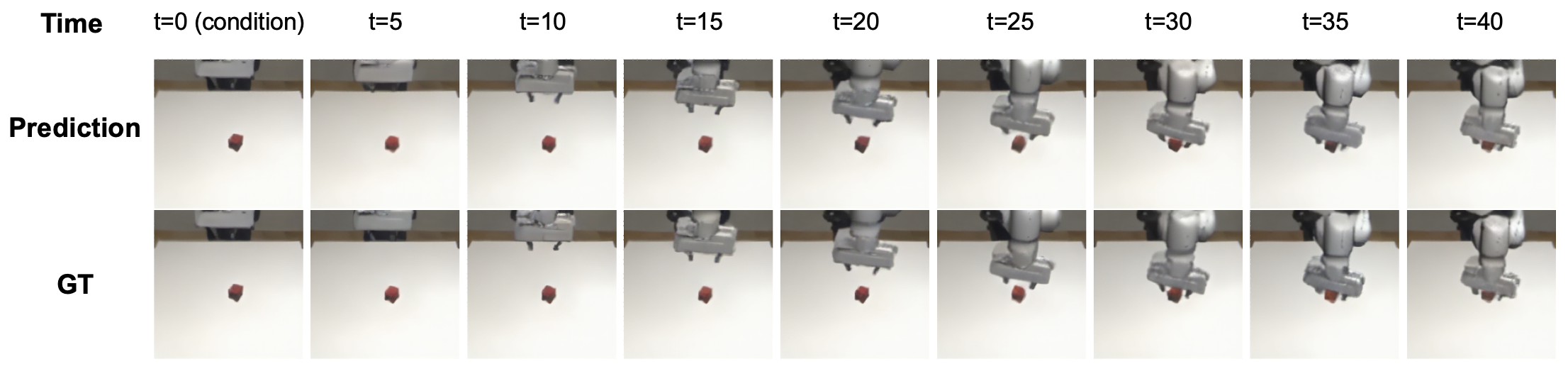}
    \caption{Supplemented visualization results on Robomimic (displaying every 5th frame; 8 frames shown from t=0 to t=40). }
    \label{fig:result_robomimic}
\end{figure}

\begin{figure}[t]
    \centering
    \includegraphics[width=\linewidth]{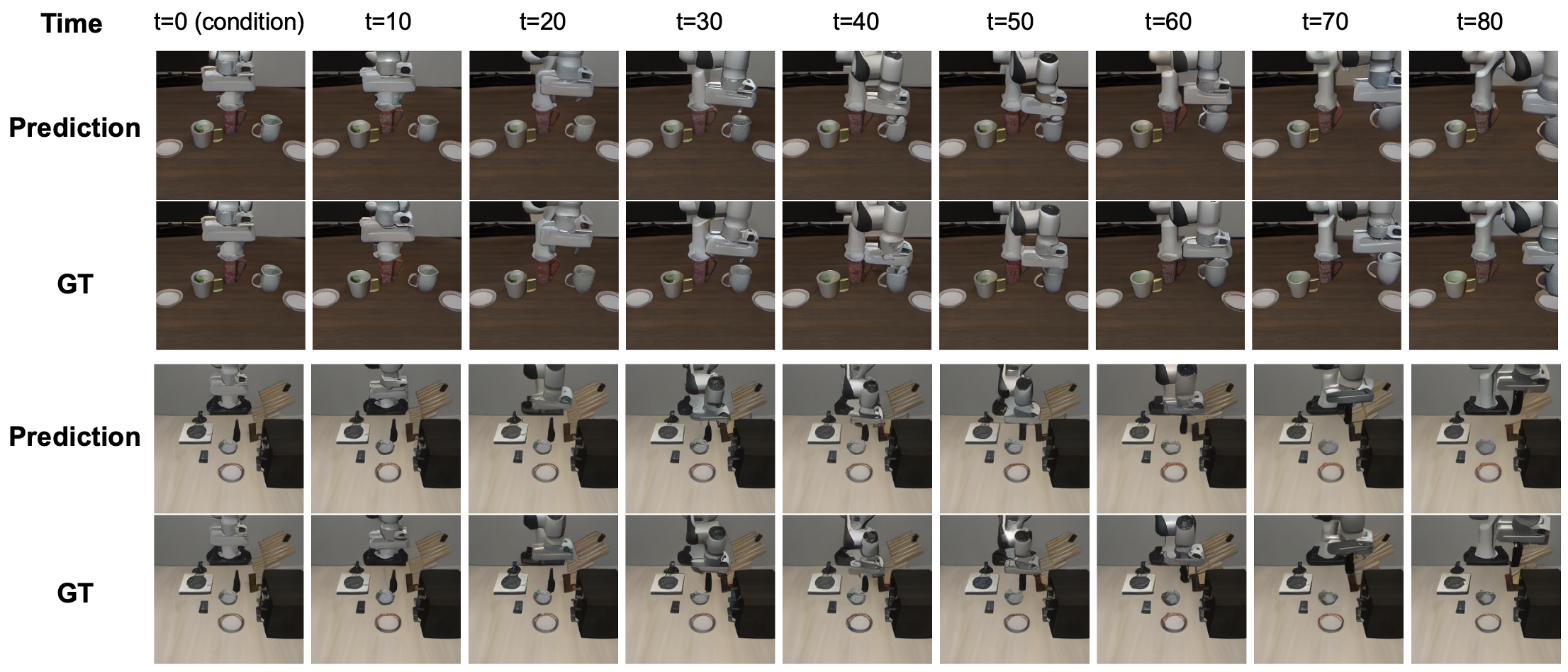}
    \caption{Supplemented visualization results on LIBERO (displaying every 10th frame; 8 frames shown from t=0 to t=80). }
    \label{fig:result_LIBERO}
\end{figure}

We provide some visualization results of generated data on Robomimic and LIBERO using our model, which are shown in Figure~\ref{fig:result_robomimic} and Figure~\ref{fig:result_LIBERO}.

\subsection{Robotic Policy Evaluation (add visualization results of our model and baselines}

\begin{figure}[t]
    \centering
    \includegraphics[width=\linewidth]{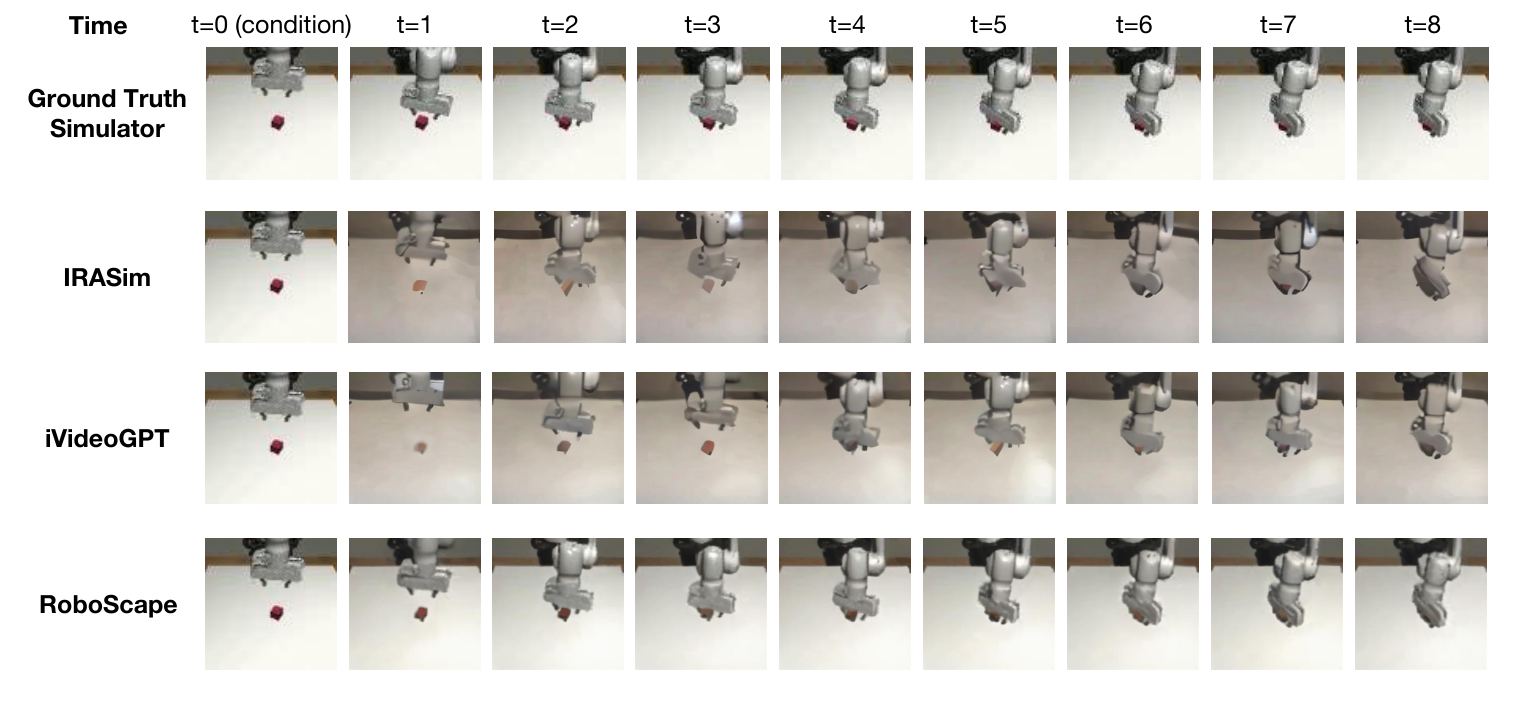}
    \caption{Supplemented visualization results of failure cases in policy evaluation. }
    \label{fig:Sup_PE_failure}
\end{figure}

\begin{figure}[t]
    \centering
    \includegraphics[width=\linewidth]{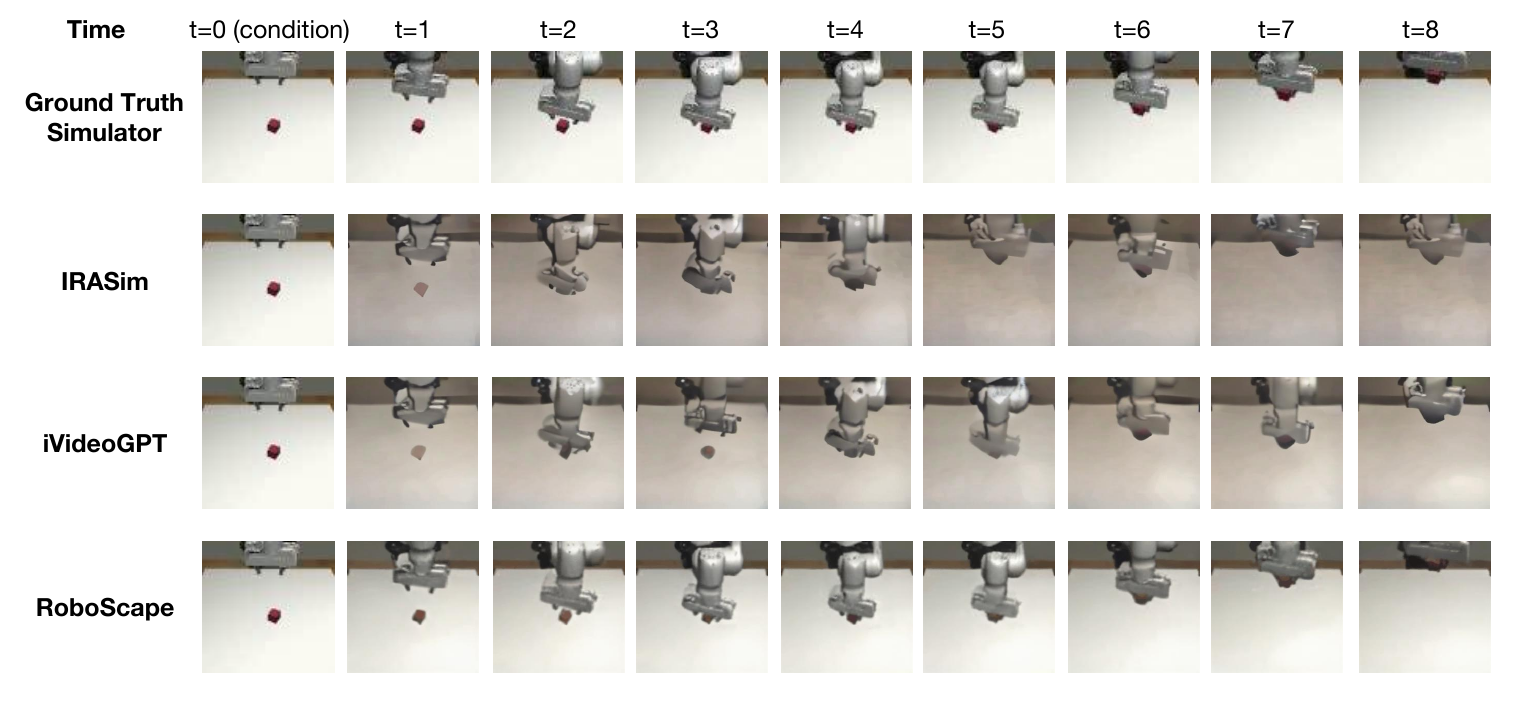}
    \caption{Supplemented visualization results of successful cases in policy evaluation. }
    \label{fig:Sup_PE_success}
\end{figure}

In this part, we provide visualization results of RoboScape and other baselines in policy evaluation. The failure cases are presented in Figure~\ref{fig:Sup_PE_failure} while the successful cases are shown in Figure~\ref{fig:Sup_PE_success}.

\end{document}